\documentclass[a4paper,conference]{IEEEtran}
\IEEEoverridecommandlockouts
\usepackage{cite}
\usepackage{amsmath,amssymb,amsfonts,array}
\usepackage{algorithmic}
\usepackage{booktabs}
\usepackage{graphicx}
\usepackage{subfigure}
\usepackage{textcomp}
\usepackage{xcolor}
\usepackage{ulem}
\def\BibTeX{{\rm B\kern-.05em{\sc i\kern-.025em b}\kern-.08em
    T\kern-.1667em\lower.7ex\hbox{E}\kern-.125emX}}
    
\newenvironment{conditions}[1][where]
    {    #1 \begin{tabular}[t]{>{$}l<{$} @{${}={}$} l}}
    {\end{tabular}\\[\belowdisplayskip]}

\newcommand{\specialcellbold}[2][c]{%
  \bfseries
  \begin{tabular}[#1]{@{}l@{}}#2\end{tabular}%
}
\begin{document}

\title{Variational Capsule Encoder\\}

\author{\IEEEauthorblockN{Harish RaviPrakash}
\IEEEauthorblockA{\textit{Department of Computer Science} \\
\textit{University of Central Florida}\\
Orlando, USA \\
0000-0002-6073-2241}
\and
\IEEEauthorblockN{Syed Muhammad Anwar}
\IEEEauthorblockA{\textit{Department of Computer Science} \\
\textit{University of Central Florida}\\
Orlando, USA \\
0000-0002-8179-3959}
\and
\IEEEauthorblockN{Ulas Bagci}
\IEEEauthorblockA{\textit{Department of Computer Science} \\
\textit{University of Central Florida}\\
Orlando, USA \\
0000-0001-7379-6829}
}

\maketitle

\begin{abstract}
We propose a novel capsule network based variational encoder architecture,
called Bayesian capsules (B-Caps), to modulate the mean and standard deviation of the sampling distribution in the latent space. We hypothesized that this approach can learn a better representation of features in the latent space than traditional approaches. Our hypothesis was tested by using the learned latent variables for image reconstruction task, where for MNIST and Fashion-MNIST datasets, different classes were separated successfully in the latent space using our proposed model. Our experimental results have shown improved reconstruction and classification performances for both datasets adding credence to our hypothesis. We also showed that by increasing the latent space dimension, the proposed B-Caps was able to learn a better representation when compared to the traditional variational auto-encoders (VAE). Hence our results indicate the strength of capsule networks in representation learning which has never been examined under the VAE settings before.
\end{abstract}


\section{Introduction}
\label{submission}
Autoencoders (AEs) have been around since the 1980s~\cite{ballard1987modular} and are used for encoding the input into a latent space that optimally represents 
high dimensional data with lower dimensions by introducing a bottleneck layer in the encoding-decoding process. 
Due to this representation ability, AEs can naturally be used to extract 
features for various classification/detection tasks~\cite{vincent2008extracting} \cite{zhou2019learning}. In a typical AE, the input data is passed through a few or several neural network layers to obtain a reduced and more compact dimensional representation. This manageable delineation, \textit{the encoding vector}, encodes the different learned attributes. Thus, the learned latent space encodes descriptive attributes of the data and can be used for several purposes including classification and reconstruction. To understand and model the variations associated with these attributes, a probabilistic distribution estimate of the latent variables can be used. In this regard, variational inference is a general way to capture variations in the data by approximating 
the probability densities and converting the inference problem into an optimization problem, which is ideally suited for machine learning/deep learning settings. Variational inference methods approximate the posterior distribution of the data and latent variables and use Kullback-Leibler (KL) divergence to measure the difference between the approximate and true posteriors~\cite{kullback1951information}. Hence, minimizing the KL divergence becomes an optimization problem.

\subsubsection*{Variational Autoencoder (VAE)}
Variational autoencoder is the simplest model for applying variational inference to deep learning based methods~\cite{kingma2013auto}. A VAE can help in establishing a meaningful relationship between the raw input data $(x)$ and the feature representation in latent space $(z)$. A VAE represents a parametric generative model $p_{\sigma}(x|z)$, with a posterior probability of the inference model $q_{\gamma}(z|x)$. Ideally, these generative and inference models should be equal, which is not the case in actual practice. Usually, $q_{\gamma}(z|x)$ is often taken to be a Gaussian distribution. This in turn could help in understanding different variations within each class of the data. For instance, for a hand written character written by different people, instead of learning a separate representation for each instance, a VAE learns the variations and approximately reconstructs (recognize) the digit. This is usually achieved by generating a continuous latent space as opposed to a conventional AE. Despite having these advantages, both AE and VAE are \textbf{not} viewpoint invariant; hence, they either require a large amount of data for precise modeling or certain bounds on the learning process for a fair representation \cite{moyer2018invariant}. In our proposed study based on capsule networks, we develop a new algorithm which could address viewpoint invariant representation and learning part-whole relationships in a variational encoder setting. In the following, after a brief background on capsule networks and related works, we present the details of the proposed B-Caps algorithm.

\subsubsection*{Capsule Networks}
Although convolutional neural networks (CNNs) have been successful in a wide spectrum of classification and detection applications, their performance could suffer when data representations have varying or novel viewpoints. These variations in the appearance manifold could likely be learned by carefully designed data augmentation methods, albeit adding to the computational cost. A traditional VAE would also fail to model these relationships in the latent space which thereby reduces the invariance of such models under various image transformations. Learning models that are transformation invariant to such manifolds has been a challenging task. Capsule networks, by replacing scalar neurons with vectors, assist in learning a relationship between objects and its parts \cite{sabour2017dynamic}. Capsules make the underlying assumption of objects or entities being composed of parts and ideally, learning the part-whole relationship for these entities benefits the learning process by making it invariant to transformations and novel viewpoints. 
Towards this end, the proposed B-Caps uses advantages of capsule networks under a Bayesian setting as described in Section~\ref{vce}.

\vspace{-3mm}
\subsection{Related Work}
Since the introduction in 2013, several variations of VAEs have been proposed to cater for different tasks and domains in representation learning. 
A deep CNN based encoder and a deep generative deconvolutional network (decoder) was proposed for modeling images and their captions \cite{pu2016variational}. The learned model was able to run in a semi-supervised setting in test cases where the labels were not available. In natural language processing, Kusner et. al.~\cite{kusner2017grammar} proposed a grammar VAE to incorporate knowledge about the structure of data and applied this model to parse trees. In~\cite{chen2016variational}, a variational lossy AE was proposed to learn more global representations while dropping local ones. The authors combined the VAE with recurrent neural networks to achieve this goal. Along similar lines,  Habibie et. al. proposed to learn the manifold of human motion from motion capture dataset using a recurrent VAE~\cite{habibie2017recurrent}. It was observed for deep stochastic models, that starting with the reconstruction loss before introducing the KL loss was important for convergence \cite{sonderby2016ladder}. It was also noted that batch normalization played an important role in these networks. A shape VAE was proposed, which modeled the distribution of object parts, locations of surface points, and the normal associated with these points~\cite{nash2017shape}. The modeling of the distribution of object parts and locations attempted to model the part-part relationships. On the contrary, capsule networks took this a level up by modeling an object-part relationship and transforming the AE into a classification network~\cite{sabour2017dynamic}. Following this, the capsule networks were applied to a range of applications from text classification~\cite{zhao2018investigating} and action detection~\cite{duarte2018videocapsulenet} to brain tumor classification~\cite{afshar2018brain} and explainable medical diagnoses~\cite{lalonde2019encoding}. 

In certain domains, where the availability of labeled data is scarce, VAEs 
worked well to support semi-supervised learning \cite{li2019disentangled}: the proposed architecture combined the latent space and reinforcement learning to enable learning for data with limited labels. For an effective inference from latent variables in generative modeling, a Bayesian approach towards learning the latent representation could play a significant role. In a recent study, a routing algorithm was proposed for capsules inspired by variational Bayes \cite{ribeiro2019capsule}. The network consisted of a convolution layer, a primary capsule layer, 2 convolutional capsule layers, and a fully connected capsule layer which was used for classification task. However, there is still little evidence of work which transforms a VAE such that the latent space representation makes use of the part-whole relationship probabilities. We argue that such a
representation could be inferred by learning a latent embedding using capsules during the encoding process. Towards this overreaching goal, we used a shallow network with fully connected capsules for the image reconstruction task. 

\subsection{Summary of Our Contributions}
We argue whether a more powerful structured representation in the inference model 
is possible with capsule networks.
To this end, we design a new VAE with the following contributions: 
\begin{itemize}
    \item We propose a novel VAE architecture, called Bayesian Capsules (or B-Caps), which combines VAE and capsule networks 
    by utilizing the variational Bayes approach. 
    \item The proposed B-Caps helps modulate the mean and standard deviation of the latent space distribution, which in turn generates improved reconstructed images when compared with a traditional baseline VAE. The results were evaluated using both the MNIST and Fashion-MNIST datasets. 
    \item The representative power of the learnt latent distribution was evaluated by learning a classifier, with a significant classification performance compared to baseline models. 
    \item There are also some incremental novelties in our study such as the use of batch normalization in capsule layers to make the learning faster and helping the network to generalize better.
\end{itemize}

\section{Methods}
Herein, we briefly outline the VAE architecture and capsule network layers before presenting our proposed fusion of these two concepts.

\subsection{Variational Autoencoder}
Unlike the vanilla AE, a VAE generates two outputs in the encoder: a vector of \textit{means} and a vector of \textit{standard deviations}. These outputs form the parameters of a vector of random variables from which latent samples are generated. This helps the encoder in learning a (potentially) different mean for each class while the standard deviation controls its spread and reduces the overlap with other classes. VAEs are typically trained using two losses - \textit{a generative loss}, which measures the accuracy of the reconstructed image, and \textit{a latent loss} which measures how closely the latent variables are distributed to a unit Gaussian. The latent distribution loss is controlled using the Kullback-Leibler (KL) divergence:
\begin{equation}
    D_{KL}(P||Q) = -\sum_{z\in\chi}P(z)\log\frac{Q(z)}{P(z)},
\end{equation}
\begin{conditions}
 \chi & probability space,\\
 P \& Q & probability distributions.
\end{conditions}
In the VAE, $P$ is the latent variable and $Q$ is a unit Gaussian. To enable random sampling and support backpropagation for optimizing the KL loss, a reparameterization trick is usually employed. The variable $z$, using the samples from the encoder outputs - including mean ($\mu$) and standard deviation ($\sigma$), can be sampled from the standard deviations such that the mean is added afterwards:
\begin{equation}\label{eq:2}
    z = \mu + \sigma\cdot\epsilon,
\end{equation}
where $\epsilon$ $\in$ $\mathcal{N}(0,1)$. A more comprehensive summary of the VAEs can be found in~\cite{kingma2019introduction}.

\subsection{Capsule network layers}
Unlike a fully connected layer, every capsule layer (in capsule networks) is either 2-dimensional or $N+1$ dimensional (convolutional capsules), where $N$ is the convolution dimension~\cite{lalonde2018capsules}. The additional dimension converts the scalar filter into a vector representation, thereby enabling the encoding of pose and orientation information. The vector length along this dimension gives the probability (that an object/object-part exists) and the orientation with respect to its parent layer (i.e. input image or previous capsule layer).
Dynamic routing is utilized to enable grouping of capsules such that 
similar lower level capsules are grouped together to a higher level capsule. Considering a lower level child capsule ($c_{i}$) and a higher level parent capsule ($c_{j}$), the description vector $u_{i}$ of $c_{i}$ is related to $c_{j}$ via transformation matrix weights $W_{ij}$, which are trained via backpropagation. Hence, the child capsule predicts the output of the parent capsule as: 
\[
    \hat{u_{j|i}}=W_{ij}u_{i}.
\]
The output $u_{j}$ of $c_{j}$ is computed as, 
\[
    u_{j} = \sum_{}k_{ij}\hat{u_{j|i}},
\]
where $k_{ij}$ is a coupling coefficient. The output description $u_{j}$ is normalized to [0, 1] via a \textit{squashing function}~\cite{sabour2017dynamic}. In this way, the description vector can be seen as the probability of detecting a feature with a given orientation. This builds the basic object (input capsule) and object-parts (child capsules/group of capsules) relationship. 
Different routing algorithms such as expectation maximization (EM) routing~\cite{sabour2018matrix}, self routing~\cite{hahn2019self}, dynamic routing with min-max normalization~\cite{zhao2019capsule}, and variational Bayes \cite{ribeiro2019capsule} have been proposed to reduce the computational burden and improve the coupling between capsules. It should be noted that exploring different routing algorithms are kept outside the scope of the current work.
\begin{figure*}[!ht]
    \centering
    \subfigure[B-Caps encoder architecture \label{fig:varCaps}]
    {{\includegraphics[width=0.45\textwidth, height=60mm]{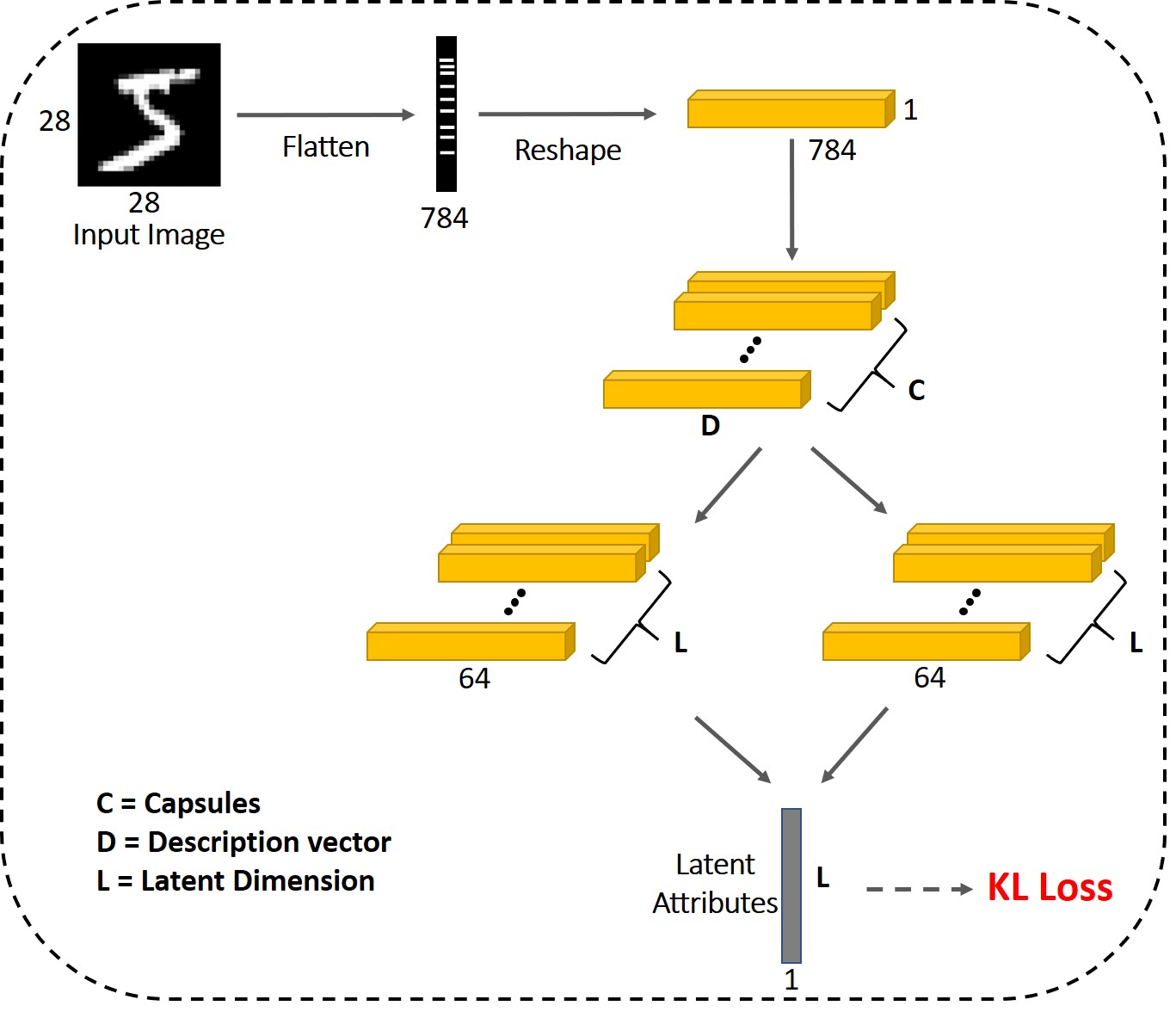} }}\hspace{-1.2em}%
    \qquad
    \subfigure[Decoder architecture \label{fig:decoder} ]{{\includegraphics[width=0.45\textwidth]{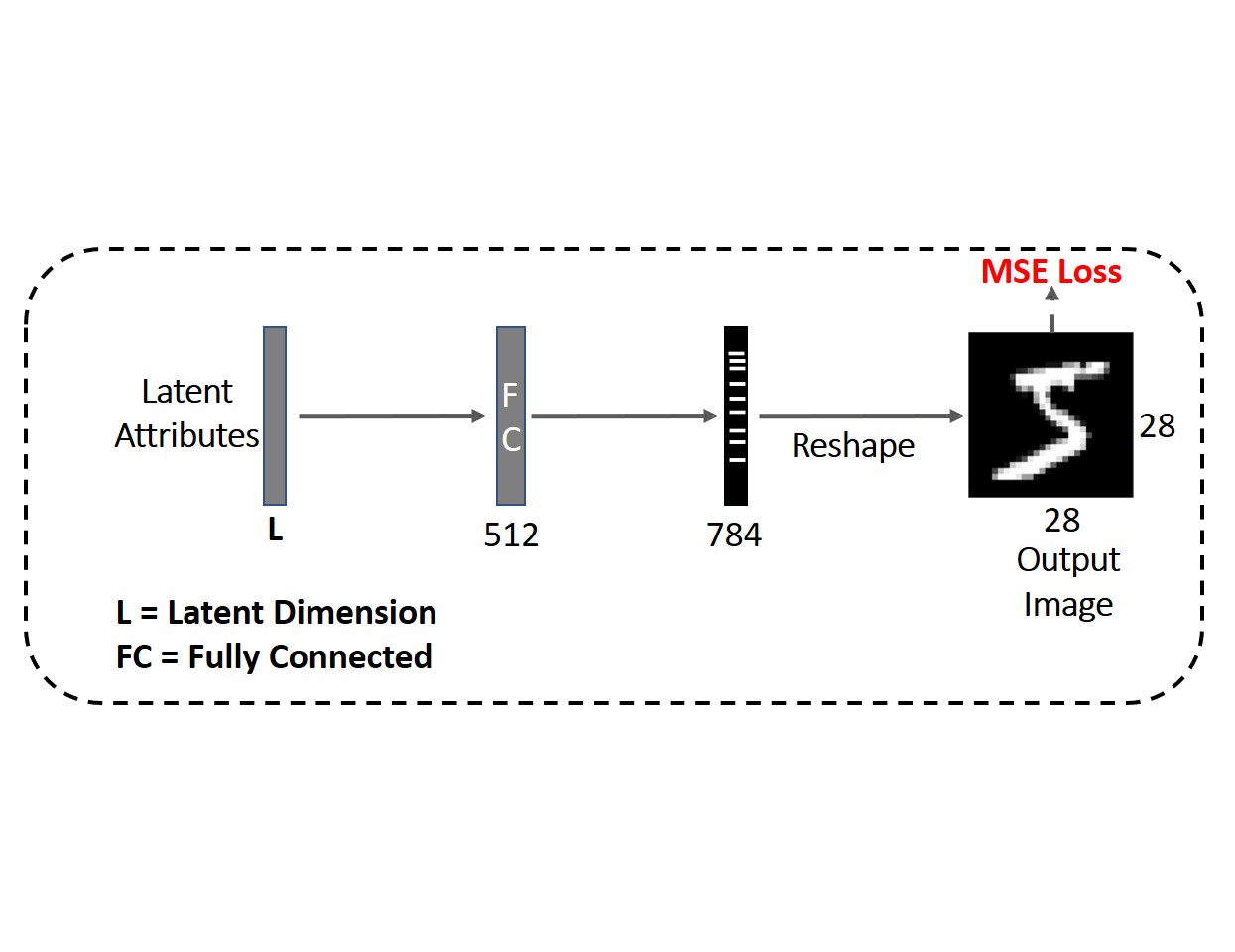} }}%
    \caption{B-Caps architecture: Represented by three parameters; the number of capsules ($C$), the size of description vector ($D$), and the latent dimension ($L$). The latent attribute vector is generated from the euclidean norm of the preceding layer.}
    \label{fig:BCaps}
\end{figure*}

\subsection{Variational Capsule Encoder}\label{vce}

In this study, we investigate the ability of capsules for learning feature variations in latent space. For this, we aim to build strong relationships between the image and the object parts even in a shallow VAE network where the encoder part includes capsule layers.
Traditionally in capsule networks, a convolutional layer is employed to generate features which are then converted into a primary capsule layer. In this way, the channels serve as the description vector of the single capsule. Herein, we propose to skip this step for small images like those in the MNIST and Fashion-MNIST datasets. Instead, we treat the flattened image as a description, thereby converting the whole image into a capsule. We posit that individual pixels are descriptive of the image and thus can be treated as a vector description for a single image capsule. 

The proposed variational capsule encoder architecture is shown in Fig. \ref{fig:varCaps}. The output consists of two groups (mean and standard deviation) of capsule layers, each with $L$ capsules. The length ($L$) of these layers defines a vector of random variables of length $L$ in the latent space. 
The network architecture is represented by three different parameters: the number of capsules (C), the size of description vector (D), and the latent dimension (L). The flattened input image is routed to $C$ capsules of vector length $D$. These are then routed to the mean and standard deviation capsules having $L$ capsules, whose vector norm defines the latent attribute vector.

Traditionally, in capsule based networks designed for classification or segmentation tasks~\cite{lalonde2019encoding}~\cite{lalonde2018capsules}, an additional reconstruction loss can be included to encourage the capsules to encode inputs' instantiation parameters (such as the pose information). 
On the contrary, we used the reconstruction loss as our primary loss (with no additional task such as classification), and a fully connected decoder network (illustrated in Fig. \ref{fig:decoder}) was implemented to reconstruct the flattened image. 
The B-Caps network is trained to minimize the mean squared error (MSE) loss and the KL loss, thereby simultaneously optimizing the latent space as well as the image reconstruction. 
To enable sampling and use of KL in backpropagation, a resampling trick is used to approximate the random normal posterior. The input to the latent sampling layer is a vector Euclidean norm of the mean and standard deviation capsules. The norm of the description vector $u$ is computed as: 
\[
u_{i} = \sqrt{u_{i1}^{i2}+u_{2}^{2}+\cdots+u_{iD}^{2}},
\]
where i $=1 \cdots L$ and D is the size of the description vector. The latent vector $z$ is sampled as in Eq.~\ref{eq:2} and the KL loss is computed as: 
\begin{equation}
     D_{KL}(\mathcal{N}(\mu,\sigma)||\mathcal{N}(0,1)) = 0.5 \times \sum_{L}(exp(\sigma)+\mu^{2}-1-\sigma).
\end{equation}

\section{Experiments}

\subsection{Baseline VAE architecture}
We devised the baseline VAE architecture with two fully connected layers in the encoder and a fully connected layer in the decoder prior to the reconstruction layer. 
The mean ($\mu$) and the standard deviation ($\sigma$) is represented by the $L$ fully connected layers in the encoder preceding the bottleneck layer. The decoder network (Fig.~\ref{fig:decoder}) takes as input the 
latent attributes and outputs the reconstructed image. In our experiments, we fixed the dimension of the fully connected layer in the encoder and decoder to 512. Batch normalization was included after the first fully connected layer, in both the encoder and decoder. Although, the baseline VAE encoder architecture is simplistic but is kept similar (in terms of layers) to the B-Caps encoder architecture (Section III-B) for a fair comparison. Moreover, the decoder used in both architectures is the same (Fig.~\ref{fig:decoder}).


\begin{figure*}[!ht]
    \centering
    \subfigure[Standard Normal Sampling]{{\includegraphics[width=0.28\textwidth, height=40mm]{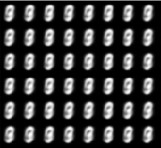}}}\hspace*{-1.2em}%
    \qquad
    \subfigure[Random Normal Sampling ]{{\includegraphics[width=0.28\textwidth, height=40mm]{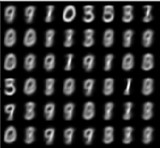} }}\hspace*{-1.5em}%
    \qquad
    \subfigure[Data-driven Sampling ]{{\includegraphics[width=0.28\textwidth, height=40mm]{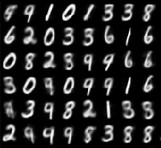} }}%
    \caption{Reconstruction results with different training strategies.}%
    \label{fig:training}
\end{figure*}

\subsection{B-Caps Architecture}
Our proposed B-Caps architecture follows the VAE formulation and has a depth of just 2 layers in the encoder and a fully connected layer in the decoder prior to the reconstruction layer. Each of these capsule layers is a fully connected capsule, also known as \textit{DigitCaps}~\cite{sabour2017dynamic}. The primary capsule layer comprises of $C$ capsule types of vector length $D$. The last layer comprises of the two outputs, the mean ($\mu_{c}$) and the standard deviation ($\sigma_{c}$) capsules, having $L$ capsule types of description length $D_{1}$. The complete variational capsule encoder is represented as \{\{$C$, $D$\},\{$L$, $D_{1}$\}\}. In our experiments we fixed $D_{1}=64$, and varied the number of capsule types in the layer prior to the output layer in the encoder part by varying $C$, hence evaluating the effect of the description length $D$. Since capsules encode part-whole relationships, the effect of increasing $C$ can change with varying $L$. This is because both the \textit{mean} and \textit{standard deviation} capsules would have $L$ capsule types and hence the part-whole relationship between $C$ and $L$ would change. Therefore, we set $L=2$ for a set of initial experiments which allowed visualizing class separation in the latent space.

\subsection{Effect of latent dimension on performance}
The quality of the reconstructed images can improve by increasing the length of the latent attributes vector \cite{kingma2013auto}. In our proposed B-Caps, the length of the latent attributes vector $L$ translates to the number of capsule types. By increasing this value, we tested the number of part-whole relations that can be learnt in a shallow network. This also enabled us to test whether learning more part-whole relationships improves the performance.

\section{Results}
\subsection{Summary of Main Results}
We trained the proposed B-Caps on MNIST and Fashion-MNIST data and observed that the model does not converge when the variance was sampled from a standard normal distribution. We hypothesized that the feature distribution in capsule layers does not follow simple Gaussian distribution; therefore, we trained B-Caps using a data driven approach to observe the distribution of data along capsule layers. We call this approach \textit{pseudo-MCMC} due to its similarities with the Markov Chain Monte Carlo (MCMC) approach. 
We also showed that with increasing dimension of the latent variables, the proposed B-Caps 
outperformed the baseline VAE, indicating that the learned attributes have a stronger relation to their preceding layers in B-Caps. 
In the following, we present our experimental results in detail.
\subsection{How to train B-Caps?}
We first trained our proposed B-Caps in a similar way as used for training a regular VAE using the standard normal distribution with the reparameterization trick. However, we noticed that backpropagation may fail and the loss function may not converge. 
A possible reason for this is that the mean and standard deviation vectors in B-Caps are driven by the length of the vectors $\mu_{c}$ and $\sigma_{c}$, respectively, which are always non-negative. Based on this observation, to better initialize the latent space sampling, we replaced the standard normal distribution with a normal distribution $\mathcal{N}$ with the following parameters: $(\mu=0.5,\sigma=0.5)$ so as to start with and learn a non-negative distribution. As seen in Fig.~\ref{fig:training}b, the trained model converged albeit to a poor reconstruction. The random normal sampling comes marginally closer to approximating the true distribution of the data. 
This also indicates that the variance cannot be sampled directly from a standard normal distribution. Based on these observations, we modified the random normal distribution to be data-driven 
by allowing the distribution to be modulated by $\mu_{c}$ and $\sigma_{c}$. Although this modulation may break the backpropagation with the absence of the independent random sampling (which the reparameterization trick entails), we can alleviate the potential of exploding gradients by using batch normalization along
the description vector dimension in the capsule layers. Since batch normalization adds a form of regularization to the network and helps accelerate the training~\cite{luo2018towards}. However, it should also be noted that since the description length varies from [0,1], there is a very low chance of a ``bad" variance in sampling that could break the backpropagation. 
In the data-driven approach, we propose to use, the samples are drawn as, 
\begin{equation}
    z = \mu_{c} + \sigma_{c}\cdot\epsilon,
\end{equation}
where $\epsilon$ = $\mathcal{N}(\mu_{c},\sigma_{c}$). This modified strategy translates to creating samples similar to each data point and repeating the process several times, like in the MCMC process. Hence, with the $\mu_{c}$ and $\sigma_{c}$ changing with every update, the latent space is learnt in an MCMC manner with the reconstruction loss being learnt via optimization. As mentioned earlier, we call this training as \textit{pseudo-MCMC}. Our experimental results have shown the data-driven approach converges with a lower network loss as compared to the random normal distribution.
\begin{figure*}[!t]
    \centering
    \subfigure[baseline VAE Latent Space]{{\includegraphics[width=0.3\textwidth,height = 45mm]{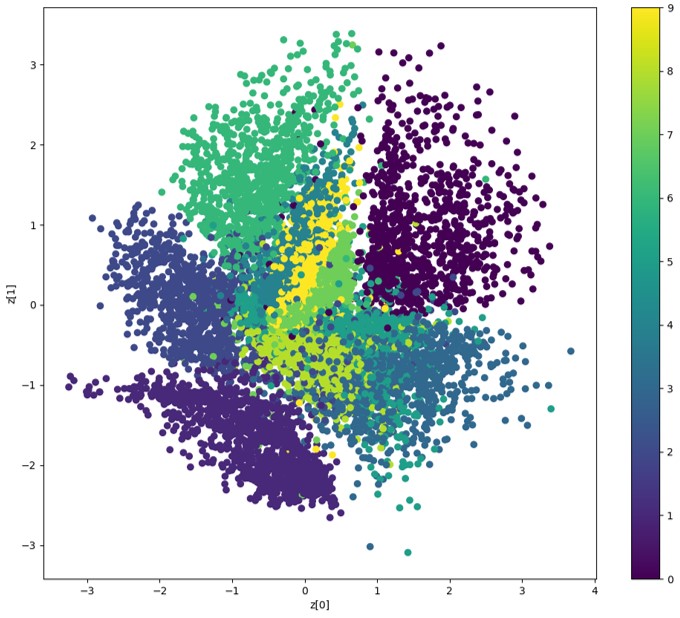}}}\hspace*{-1.2em}%
    \qquad
    \subfigure[B-Caps Latent Space ]{{\includegraphics[width=0.3\textwidth,height = 45mm]{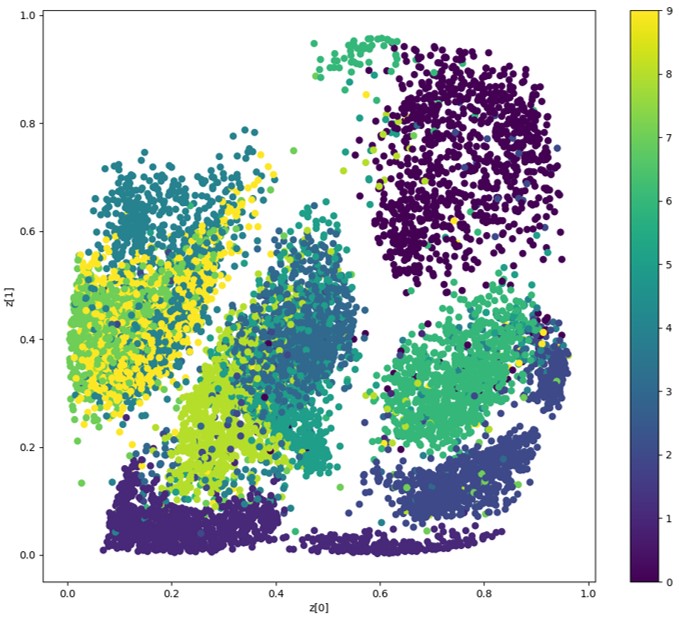} }}%
    \caption{The distribution of digits in the latent space for baseline VAE and B-Caps (C=16, D=64) on MNIST dataset. Although none of these distributions are optimal, B-Caps latent space has visibly better allocations of certain classes.}%
    \label{fig:latentspace}%
\end{figure*}

\subsection{MNIST Reconstruction}
The MNIST dataset consists of images representing hand-written digits ($0-9$) with dimension $28\times28$~\cite{deng2012mnist}. The data comprises of $60,000$ training and $10,000$ test images. We normalized all images within the $[0,1]$ range 
and flattened them before 
feeding into the B-Caps network. 

In all our experiments, the networks were trained with a batch size of $128$ for $100$ epochs. Adam optimizer was used with an initial learning rate of $1e^{-3}$~\cite{kingma2014adam}. 
During the testing phase, image reconstruction quality was evaluated using the mean squared error and the structural similarity index metric (SSIM). MSE is the mean difference between the squared pixel-wise errors between the original image and the model estimate (reconstruction). It is computed as: 
\begin{equation}
    MSE(A,B) = \frac{1}{N}\sum_{i=1}^{N}(A_{i}-B_{i})^{2},
\end{equation}
\begin{conditions}
 A,B     &  actual image, reconstructed image, \\
 N     &  number of pixels.
\end{conditions}

SSIM on the other hand, is a perceptual image quality metric assessing the effect of luminance, contrast, and image structure. SSIM is computed as a product of the aforementioned variables as: 
\begin{equation}
    SSIM(A,B) = \frac{(2\mu_{A}\mu_{B}+C_{1})(2\sigma_{AB}+C_{2})}{(\mu_{A}^{2}+\mu_{B}^{2}+C_{1})(\sigma_{A}^{2}+\sigma_{Y}^{2}+C_{2})},
\end{equation}
\begin{conditions}
 A,B     &  actual image, reconstructed image, \\
 \mu_{A}, \mu_{B}     &  means of A \& B, \\
 \sigma_{A}, \sigma_{B}     &  standard deviations of A \& B, \\
\sigma_{AB} &  cross-variance of A \& B, \\
C_{1}, C_{2} & constants to avoid instability.
\end{conditions}

\begin{table}[!h]
    \caption{Comparison of reconstruction quality on MNIST while varying the capsule types ($C$) and description length ($D$). std - standard deviation.}
    \begin{center}
    \resizebox{\columnwidth}{!}{%
    \begin{tabular}{|c|c|c|c|c|}
        \hline
        \hline
         \textbf{Model} & \specialcellbold{Capsule\\types (C)} & \specialcellbold{Description\\length (D)} & \specialcellbold{SSIM\\mean$\pm$std} & \specialcellbold{MSE\\mean$\pm$std}\\
         \hline
         \midrule
         Baseline VAE & $-$ & $-$ & $0.555\pm0.154$ & $0.041\pm0.019$\\
         B-Caps & 8 & 64 & $0.541\pm0.144$ & $0.043\pm0.020$\\
         \textbf{B-Caps} & \textbf{16} & \textbf{64} & $\textbf{0.580}\pm\textbf{0.133}$ & $\textbf{0.040}\pm\textbf{0.017}$\\
         B-Caps & 32 & 64 & $0.573\pm0.147$ & $0.041\pm0.019$ \\
         B-Caps & 8 & 128 & $0.529\pm0.152$ & $0.046\pm0.020$ \\
         B-Caps & 16 & 128 & $0.577\pm0.129$ & $0.040\pm0.018$\\
         \hline
    \end{tabular}}
    \label{tab:tab1}
    \end{center}

\end{table}
We compared the baseline VAE at a latent dimension of $L=2$ against various configurations of the B-Caps architecture. For B-Caps, we first fixed the value of description length ($D$) and varied capsule types ($C$). Later, we increased the value of $D$ and repeated the experiments by varying $C$. We observed 
that too many capsule types in the intermediate layers 
and a larger description length ($D$) do not help in improving the reconstruction (see Table~\ref{tab:tab1}). 
Compared to the baseline VAE, at latent dimension $L=2$, B-Caps \{\{$16$, $64$\},\{$2$, $64$\}\} performs better. We visualized the latent space in Fig.~\ref{fig:latentspace}: as illustrated, while the baseline VAE gets close to the standard normal distribution with different classes radiating outwards in the 2D space, the latent space for B-Caps shows a clear separation for most of the classes.

\begin{figure}[!t]
    \centering
    \includegraphics[width=0.9\columnwidth, height = 45mm]{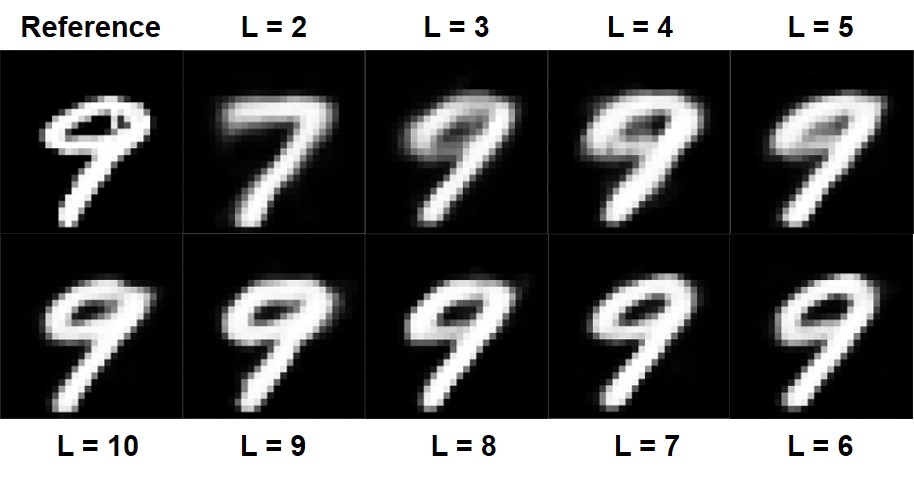}
    \caption{Improvement in reconstruction of MNIST digits as a function of the latent variable length (L).}\vspace{-1.5em} 
    \label{fig:mnist_recon}
\end{figure}

As mentioned earlier, increasing the number of latent variables improves performance when using the baseline VAE architecture. We tested how this fact can be translated into B-Caps. First, we chose B-Caps \{\{$8$, $64$\},\{$L$, $64$\}\} as our base model and varied $L$ from $2\rightarrow10$. 
The effect of varying latent dimension is seen in Fig.~\ref{fig:mnist_recon}, where we observed a poor initial guess, which started to improve at higher dimensions with performance superior to that of baseline VAE. 
We also compared the effect of latent dimension on image reconstruction quality for our proposed B-Caps and baseline VAE (Fig. \ref{fig:ssimPerf}). We observed that the B-Caps network 
outperformed the baseline with increasing $L$ and that the improvement was more discernible beyond a latent dimension of $L=4$. This adds credence to our hypothesis that the latent representation (in higher dimensions) of B-Caps is more powerful than conventional VAEs. 

\subsection{Fashion-MNIST Reconstruction}
Fashion-MNIST dataset is similar to MNIST, having images of dimension $28\times28$ split into 10 classes related to fashion products~\cite{xiao2017fashion}. There are $60,000$ training images and $10,000$ test images. Similar to our experiments for MNIST, we normalized the data to range between $[0,1]$ and flattened images to a vector of length $784$.
We evalauted different B-Caps architectures as well as the baseline VAE at a latent dimension $L=2$ (similar to MNIST experiments). 
The baseline VAE performed marginally better than the B-Caps architectures at lower latent dimension, 
indicating that the number of capsule types is more important than the description length. 

\begin{figure}[!t]
    \centering
    \includegraphics[width=0.9\columnwidth, height= 45mm]{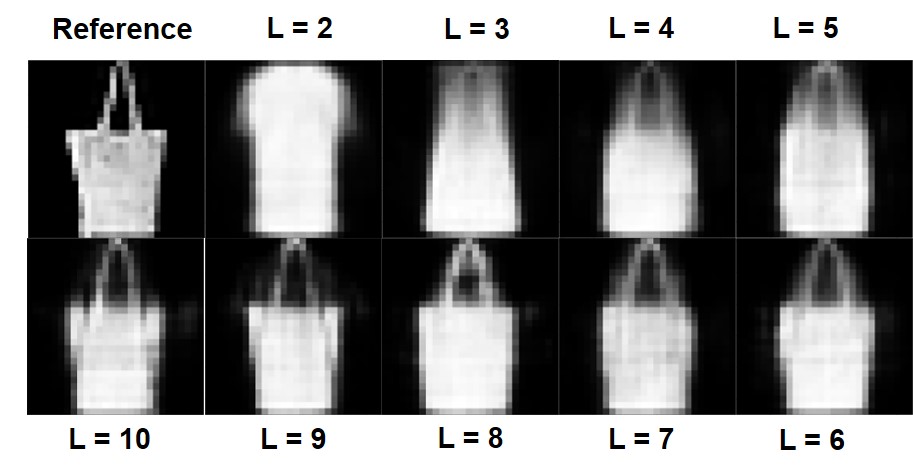}
    \caption{Improvement in reconstruction of Fashion-MNIST `bag' class as a function of the latent variable length (L).}\vspace{-1.5em} 
    \label{fig:fmnist_bag}
\end{figure}

The performance of our proposed B-Caps models was evaluated with varying dimension of the latent space. We obtained an improvement in the performance with an increase in the latent space dimension (see Fig.~\ref{fig:fmnist_bag}). 

\begin{figure}[!t]
    \centering
    \includegraphics[width=0.85\columnwidth, height = 40mm]{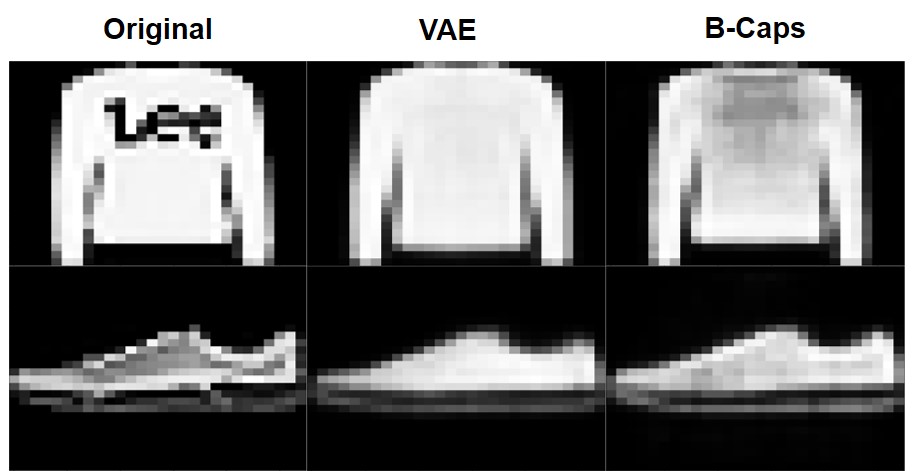}
    \caption{A comparison between reconstruction of Fashion-MNIST images using baseline VAE and B-Caps. Left to Right: Original image, Baseline VAE reconstruction and B-Caps reconstruction for $L=10$.}\vspace{-1.0em}%
    \label{fig:fmnist_compare}
\end{figure}

\begin{figure}[!ht]%
    \centering
    \includegraphics[width=\columnwidth,height=50mm]{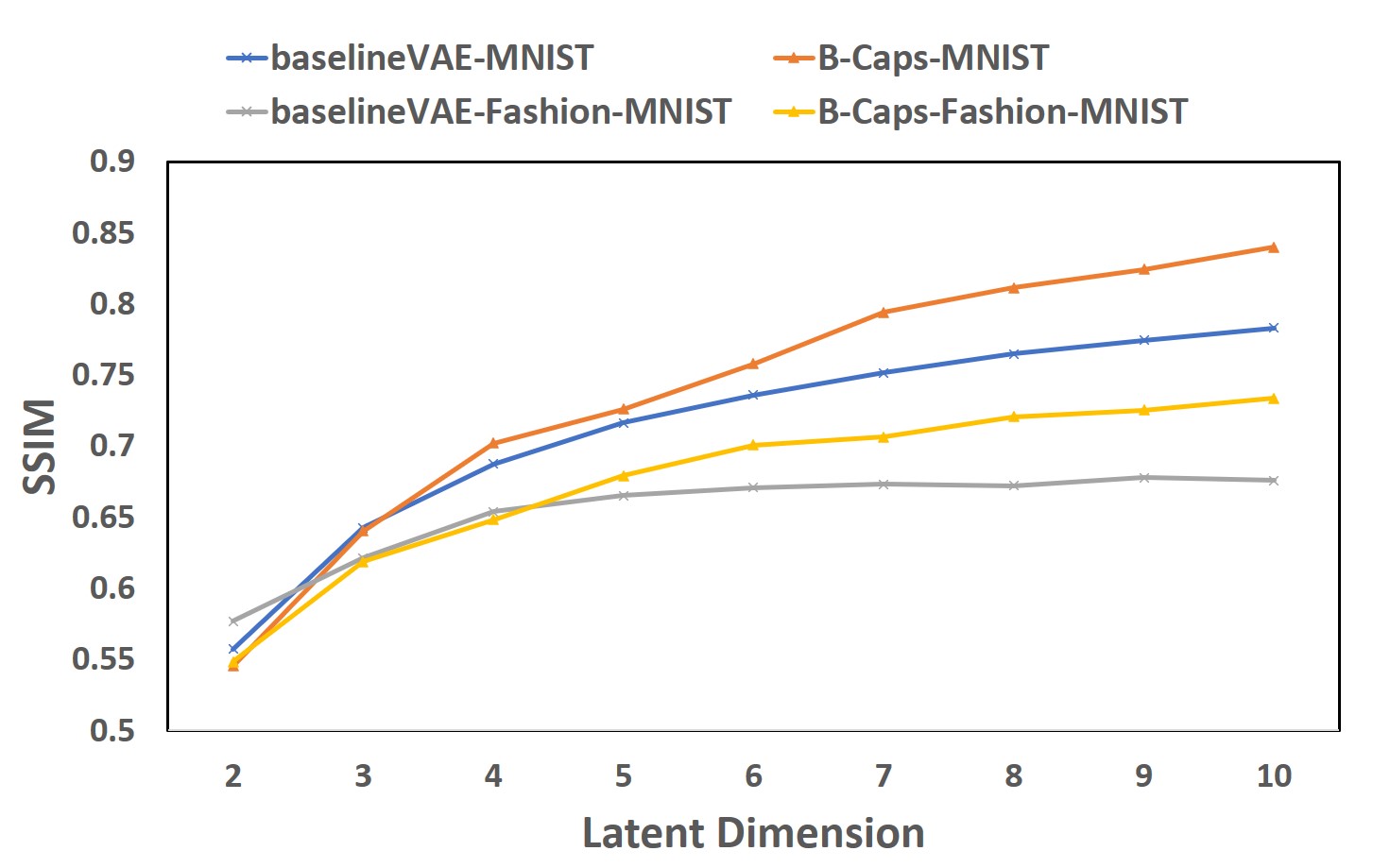}
    \caption{Variation in image reconstruction quality measured using SSIM for different latent variable dimensions.}%
    \label{fig:ssimPerf}%
\end{figure}

We compared the reconstruction of the model \{\{$8$, $64$\},\{$2$, $64$\}\} against the baseline VAE for latent dimension $L=10$. Fig. \ref{fig:fmnist_compare} shows this comparison for two different clothing categories. While both approaches capture the overall shape of the objects, B-Caps appears to capture more of the texture information within these images. For Fashion-MNIST, in our qualitative evaluations, B-Caps showed better reconstruction quality when the 
latent variables dimension was higher and this trend is similar to what we obtained for MNIST (Fig. \ref{fig:ssimPerf}). We observed that while the baseline VAE plateaus around $L=5$, the performance of B-Caps (in terms of SSIM) continues to improve.


\begin{figure}[!t]
    \centering
    \includegraphics[width=\columnwidth, height=50mm]{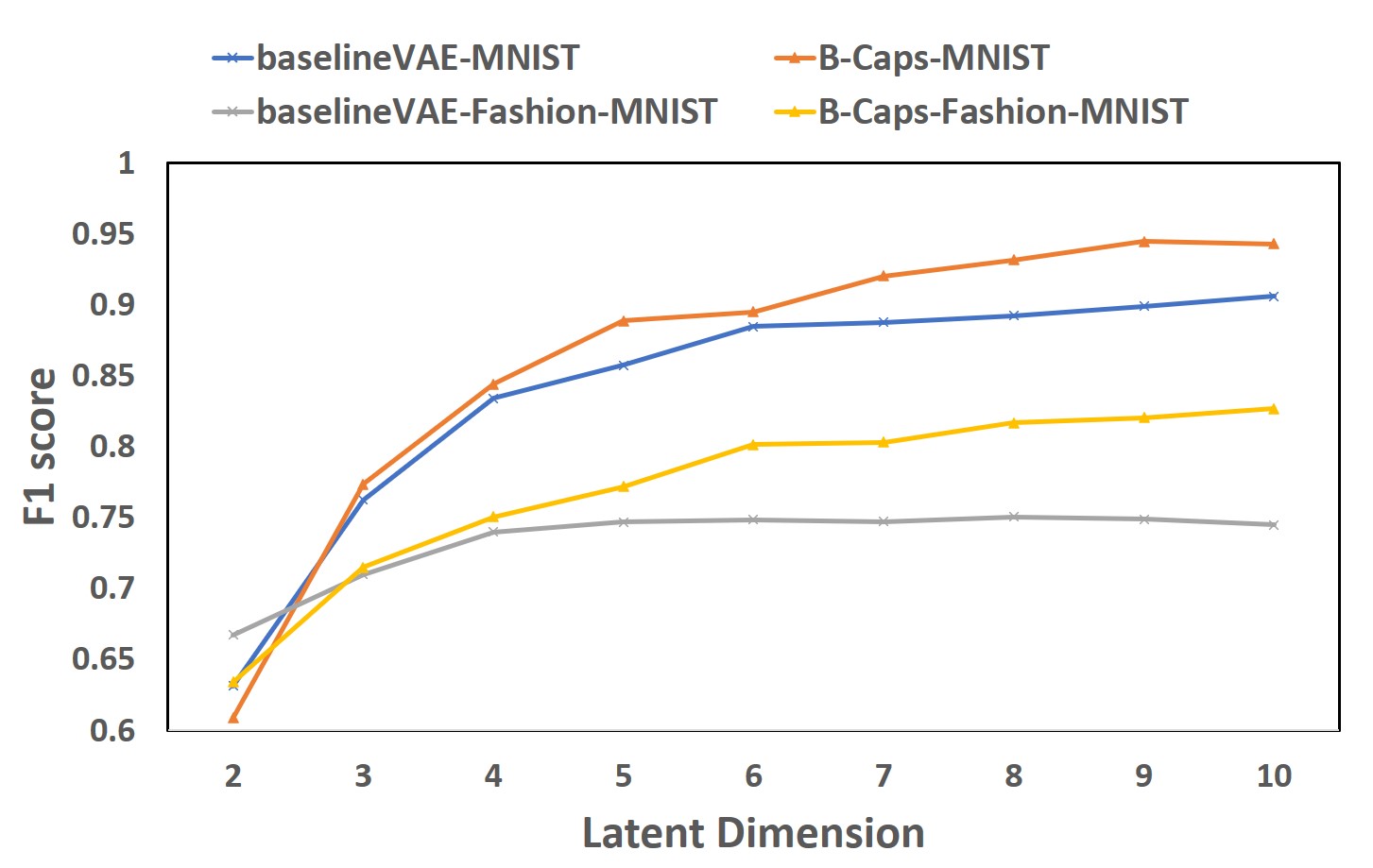}
    \caption{Variation of classification performance measured using F1 score for different latent variable dimensions.}%
    \label{fig:classPerf}%
\end{figure}
\subsection{Classification Performance}
We evaluated the performance of B-Caps in a classification setting using the reconstructed images. F1-score, the measure of a test's accuracy was used to assess the classification performance. F1-score is conventionally computed as: 
\begin{equation}
    F1 = 2\cdot\frac{precision\times recall}{precision+recall}.
\end{equation}
We trained a support vector machines (SVM) based classifier on the MNIST and Fashion-MNIST datasets. A grid search was performed to identify the best settings for the SVM for each of these datasets. For both datasets, the radial basis function (RBF) was used with a kernel coefficient of $\gamma=0.01$. The regularization parameter $K$ was set to $100$ and $10$ for MNIST and Fashion-MNIST, respectively. Once trained, the reconstructed images were evaluated 
in a classification setting. The proposed B-Caps model \{\{$8$, $64$\},\{$L$, $64$\}\} and baseline VAE, with $L$ from $2\rightarrow10$ were evaluated comparatively. Fig.~\ref{fig:classPerf} shows the variation of F1-scores with respect to the latent dimension. For both MNIST and Fashion-MNIST classification, B-Caps outperformed the baseline VAEs for $L\ge3$. Further, we can see that in Fashion-MNIST, the performance improvement is greater, indicating that B-Caps was able to capture more 
variations than the VAE.

\begin{figure}[t]
    \centering
    {\includegraphics[width=\columnwidth,height=50mm]{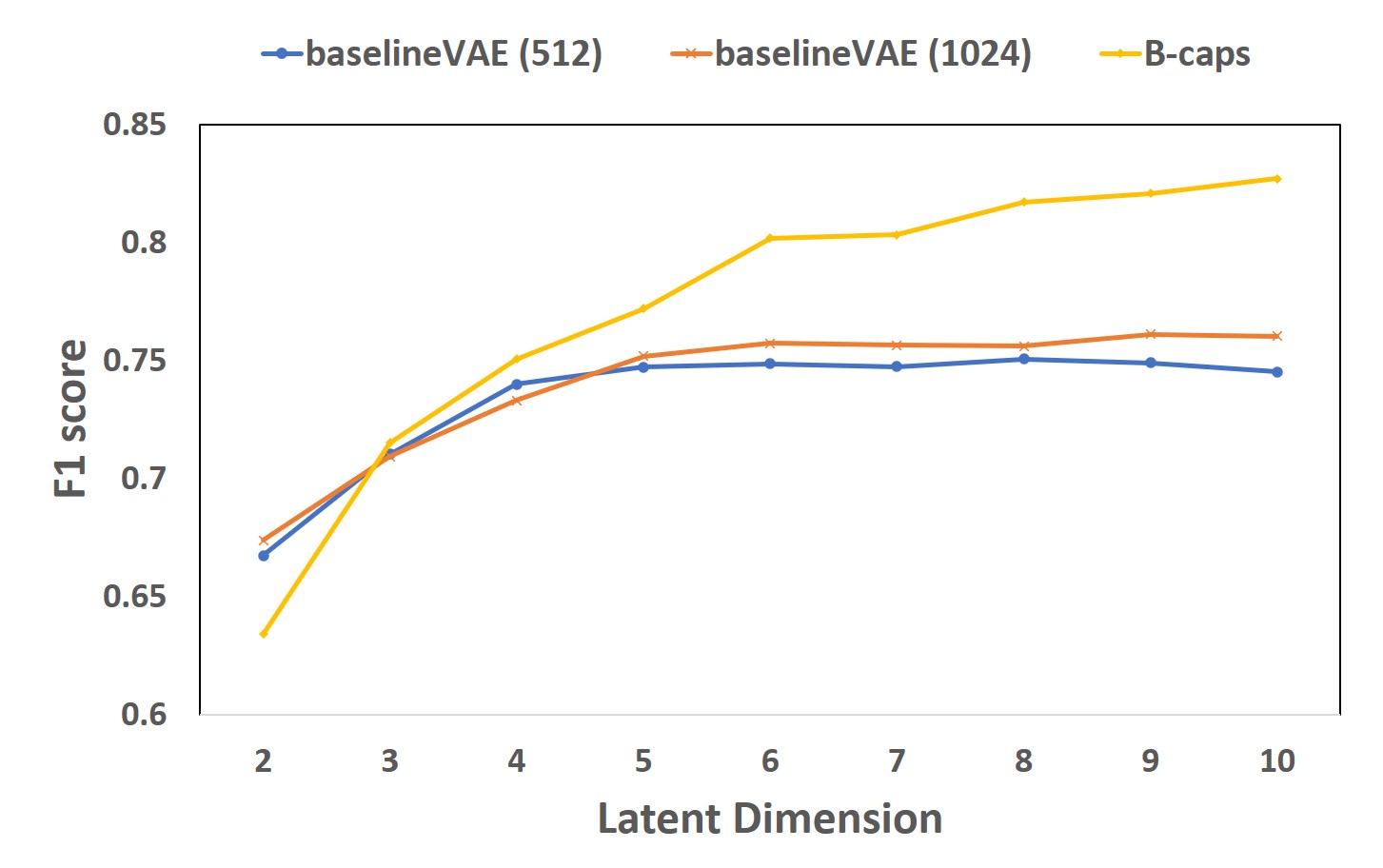}}
    \caption{Variation of classification performance with latent variable dimensions where the baselineVAE has intermediate layer of dimension 512 and 1024.}%
    \label{fig:classPerf2}%
\end{figure}

One needs to note that, an increase in the length of the latent dimension resulted in an increase in parameters for B-Caps encoder by $\approx60,000$ (per latent variable). Whereas, for baseline VAE the parameters increase by $\approx1,000$ per latent variable. To account for the difference in parameter space, we replaced the intermediate layer in the baseline VAE with a $1024$ dimension fully connected layer. The resulting comparison of the trainable encoder parameters is shown in Table~\ref{tab:tab3}. With a more comparable number of trainable network parameters, we 
repeated the comparison of the classification performance to check whether this increase in parameters for the baseline VAE would help change the plateauing effect. 
Based on the results from this new experimental setting, we conclude the overall performance was not significantly effected by increasing parameters for the baseline VAE model (see Fig.~\ref{fig:classPerf2}). 
These results revealed that the total number of parameters was not the reason for B-Caps performing better than baseline VAE. 
Instead, it is because B-Caps learns richer attributes (in latent space) owing to 
capsule layers used in the encoding process. The data driven sampling from the latent space also augmented this learning process with a better inference during the reconstruction process.  

\begin{table}[!h]
\caption{Trainable encoder parameters in the baseline VAE with intermediate layers of 512 and 1024 and B-Caps. FC- fully connected layer.}
    \begin{center}
        
   \begin{tabular}{|c|c|c|c|}
        \hline
         \specialcellbold{Latent \\Dimension} & \specialcellbold{baseline VAE \\FC-512} & \specialcellbold{baseline VAE \\FC-1024} & \specialcellbold{B-Caps\\C=8, D=64} \\
         \hline
         \midrule
         2 & 405K & 810K & 532K \\
         4 & 407K & 814K & 663K \\
         6 & 409K & 818K & 794K\\
         8 & 411K & 822K & 925K\\
         10 & 413K & 826K & 1.05M\\
         \hline
    \end{tabular}
    \label{tab:tab3}
    \end{center}
\end{table}
\vspace{-1em}

\section{Discussion and Conclusion}
We have presented a capsule based variational autoencoder architecture, called B-Caps, for an effective representation learning in the latent space and compared its performance in the image reconstruction and classification tasks. We observed the following: 
\begin{itemize}
    \item B-Caps outperformed the baseline VAE with increasing dimension of the latent space in both reconstruction quality and classification tasks.
    \item B-Caps has superiority in learning latent attributes in more complex datasets such as Fashion-MNIST compared to the baseline VAE based on the greater improvement in performance.
    \item The number of capsule types is more important than the description length.
    \item B-Caps outperformed the baseline VAE with increasing dimension of the latent space.
    \item Data-driven sampling of the latent space works better than the standard normal distribution approach in a shallow capsule encoder network. 
\end{itemize}

The reparameterization trick using an independent normal distribution enabled backpropagation and attempts to address the problem of a ``bad" variance estimate. However, in capsules, the variance is always between [0,1] as the length of the vector represents the probability of coupling between capsules. This helps in handling the ``bad" variance estimate problem but suffers from an initialization problem. The data-driven modulation of the latent space helped in handling this issue and stabilized the training. The shallowness of the proposed networks also enabled the data-driven approach. For a successful translation of the proposed approach towards deeper networks with limited data, identification of the right distribution for capsule features is significant. We have shown that this distribution cannot be a unit normal distribution. Towards this, Ribeiro et. al. recently identified that the capsule layers have a Gaussian-Wishart distribution\cite{ribeiro2019capsule} and used this to propose a novel routing algorithm. We hypothesize that in deeper networks, the data-driven approach will not be needed and a standard reparameterization trick can be used with the distribution being sampled from a Normal Wishart distribution.

It is interesting to note that the systematic selection of latent dimension is also supported by the winning lottery ticket theory, a recent study where it is shown that with the right initialization a highly pruned network could perform similar to a very dense network \cite{frankle2018lottery}. Our proposed B-Caps architecture, although not learning a sparse network, is aimed towards learning a powerful representation in the latent space. This learning is conditioned in a manner to account for the part-whole relationship (using Capsules) and embed Bayesian inference (data driven sampling from the latent space). 

In future work, we will expand upon this proof-of-concept capsule encoder by incorporating the Gaussian-Wishart distribution and further 
build convolutional capsule encoders to handle images with higher dimensions and deeper networks. We will also explore the robustness of the B-Caps under varying settings including adversarial attacks. 
\section*{Acknowledgment}
This project is partially supported by NIH R01-CA246704 and R01-CA240639, and Florida Department of Health 20K04.
\vspace{-4mm}
\bibliography{ref}

\bibliographystyle{IEEETran}
\end{document}


\title{Variational Capsule Encoder\\}

\author{\IEEEauthorblockN{Harish RaviPrakash}
\IEEEauthorblockA{\textit{Department of Computer Science} \\
\textit{University of Central Florida}\\
Orlando, USA \\
0000-0002-6073-2241}
\and
\IEEEauthorblockN{Syed Muhammad Anwar}
\IEEEauthorblockA{\textit{Department of Computer Science} \\
\textit{University of Central Florida}\\
Orlando, USA \\
0000-0002-8179-3959}
\and
\IEEEauthorblockN{Ulas Bagci}
\IEEEauthorblockA{\textit{Department of Computer Science} \\
\textit{University of Central Florida}\\
Orlando, USA \\
0000-0001-7379-6829}
}

\maketitle

\section{Results}
\subsection{Reconstruction Quality}
We compared the reconstruction quality of the baseline VAE to the proposed B-Caps at increasing latent dimensions using the SSIM metric. As seen in Table \ref{table1}, on the MNIST dataset, B-Caps outperforms the baseline VAE for latent dimensions of 4 and beyond with a comparable or lower variation in reconstruction quality. Similar results are observed for the FASHION-MNIST dataset with a consistently lower standard deviation (Table \ref{table2}).
\begin{table}[!h]
\caption{Comparison of reconstruction quality on MNIST while varying the latent dimension $L$. std - standard deviation.}
\begin{center}
    \resizebox{\columnwidth}{!}{%
\begin{tabular}{|c|c|c|}
\hline
\textbf{\begin{tabular}[c]{@{}c@{}}Latent \\ Dimension(L)\end{tabular}} & \textbf{\begin{tabular}[c]{@{}c@{}}baseline VAE\\ mean ± std\end{tabular}} & \textbf{\begin{tabular}[c]{@{}c@{}}B-Caps\\ mean ± std\end{tabular}} \\ \hline\hline
2                                                                    & 0.56 ± 0.15                                                            & 0.54 ± 0.15                                                          \\
3                                                                    & 0.64 ± 0.14                                                            & 0.64 ± 0.14                                                          \\
4                                                                    & 0.69 ± 0.13                                                            & 0.70 ± 0.12                                                          \\
5                                                                    & 0.72 ± 0.12                                                            & 0.73 ± 0.12                                                          \\
6                                                                    & 0.74 ± 0.12                                                            & 0.76 ± 0.11                                                          \\
7                                                                    & 0.75 ± 0.11                                                            & 0.79 ± 0.10                                                          \\
8                                                                    & 0.76 ± 0.10                                                            & 0.81 ± 0.09                                                          \\
9                                                                    & 0.77 ± 0.10                                                            & 0.82 ± 0.09                                                          \\
10                                                                   & 0.78 ± 0.09                                                            & \textbf{0.84 ± 0.08}   \\\hline                                             
\end{tabular}}
\label{table1}
\end{center}
\end{table}

\begin{table}[!h]
\caption{Comparison of reconstruction quality on FASHION-MNIST while varying the latent dimension $L$. std - standard deviation.}
\begin{center}
    \resizebox{\columnwidth}{!}{%
\begin{tabular}{|c|c|c|}
\hline
\textbf{\begin{tabular}[c]{@{}c@{}}Latent \\ Dimension (L)\end{tabular}} & \textbf{\begin{tabular}[c]{@{}c@{}}baseline VAE\\ mean ± std\end{tabular}} & \textbf{\begin{tabular}[c]{@{}c@{}}B-Caps\\ mean ± std\end{tabular}} \\ \hline\hline
2                                                                    & 0.58 ± 0.16                                                            & 0.55 ± 0.14                                                          \\
3                                                                    & 0.62 ± 0.16                                                            & 0.62 ± 0.15                                                          \\
4                                                                    & 0.65 ± 0.16                                                            & 0.65 ± 0.15                                                          \\
5                                                                    & 0.66 ± 0.16                                                            & 0.68 ± 0.15                                                          \\
6                                                                    & 0.67 ± 0.16                                                            & 0.70 ± 0.14                                                          \\
7                                                                    & 0.67 ± 0.16                                                            & 0.71 ± 0.15                                                          \\
8                                                                    & 0.67 ± 0.16                                                            & 0.72 ± 0.15                                                          \\
9                                                                    & 0.67 ± 0.16                                                            & 0.72 ± 0.14                                                          \\
\textbf{10}                                                                   & 0.68 ± 0.16                                                            & \textbf{0.73 ± 0.14}   \\\hline                                             
\end{tabular}}
\label{table2}
\end{center}
\end{table}
\subsection{Classification Performance}
\begin{figure*}[h]
    \centering
    \includegraphics[width=\textwidth]{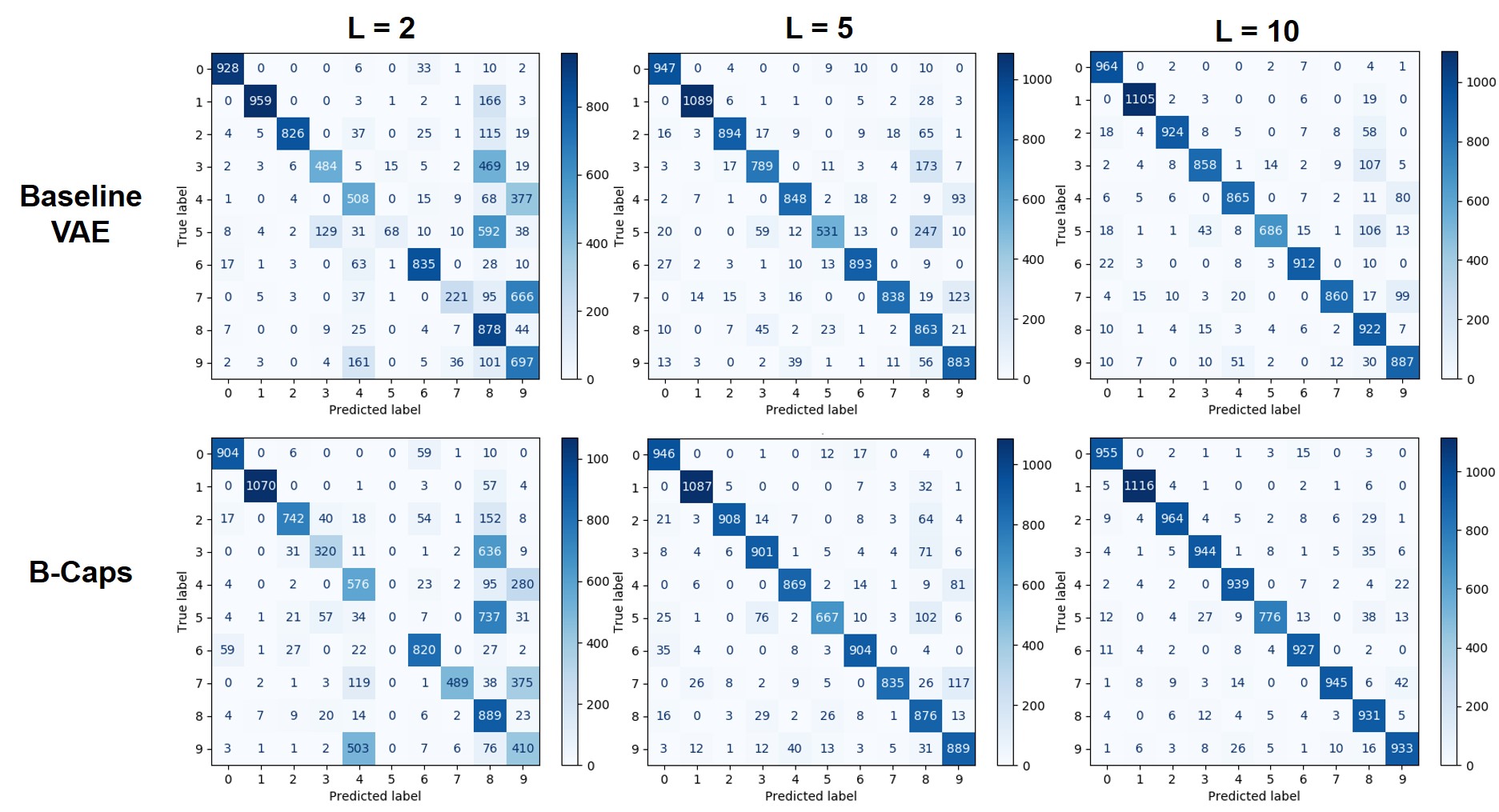}
    \caption{Confusion matrices comparing the performance of baseline VAE (top row) and B-Caps (bottom row) with different latent variable dimensions on the MNIST dataset.}%
    \label{fig:conf_mnist}%
\end{figure*}

\begin{figure*}[h]
    \centering
    \includegraphics[width=\textwidth]{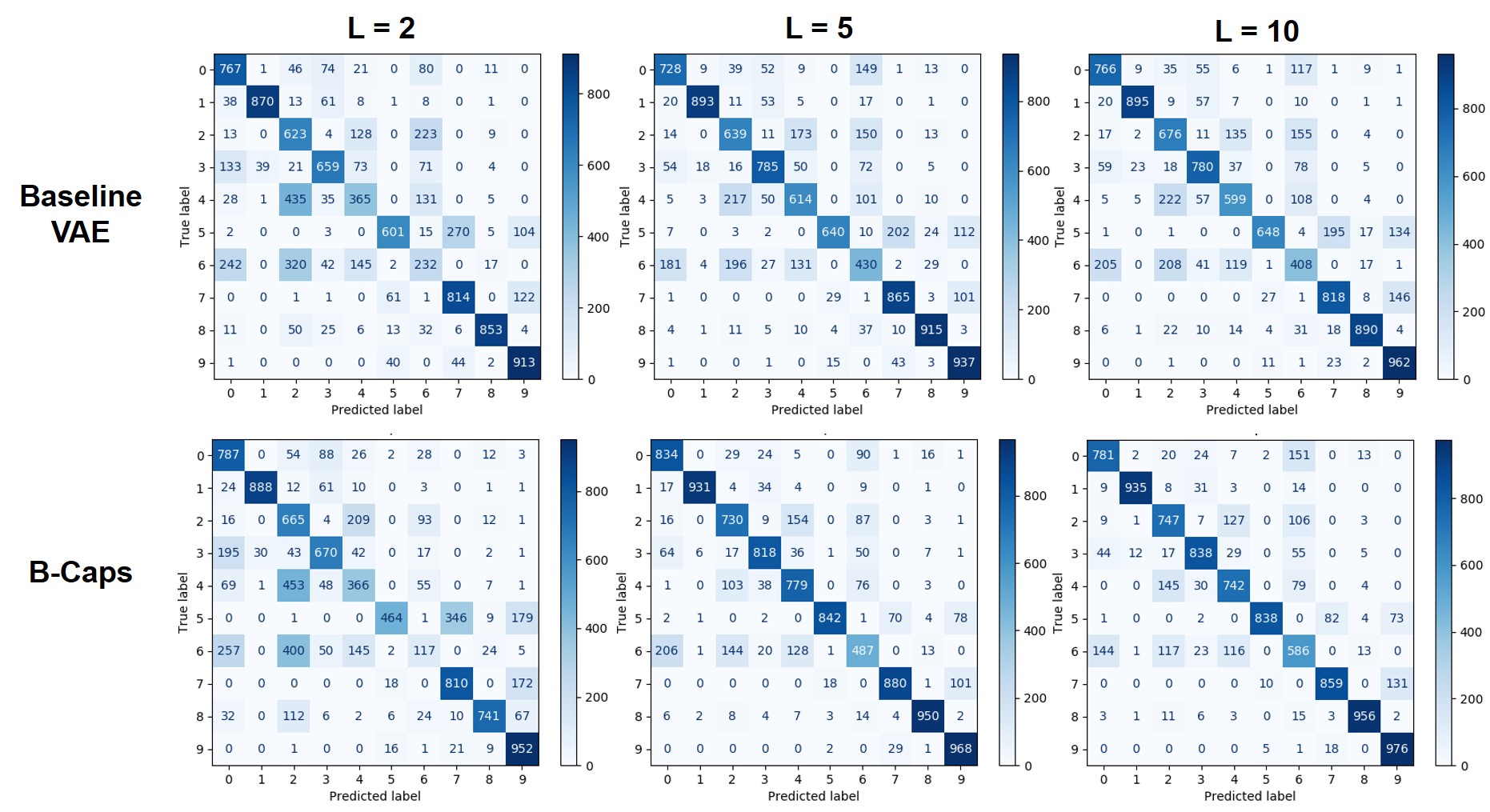}
    \caption{Confusion matrices comparing the performance of baseline VAE (top row) and B-Caps (bottom row) with different latent variable dimensions on the FASHION-MNIST dataset.}%
    \label{fig:conf_fmnist}%
\end{figure*}

Based on the results observed in the Fig.8 (main paper), we analyzed the classes influencing the improved performance in the B-Caps. On the MNIST dataset, we observe fewer misclassifications of digits 2 and 4 and significantly improved classification of the digit 7 (Fig. \ref{fig:conf_mnist}). On the FASHION-MNIST dataset we find that B-Caps has higher performance on the 'coat' class (class 4) and shirt class (class 6) (Fig. \ref{fig:conf_fmnist}). 

